\documentclass{bmvc2k}

\title{Deep Clustering by \\ Semantic Contrastive Learning}

\addauthor{Jiabo Huang}{jiabo.huang@qmul.ac.uk}{1}
\addauthor{Shaogang Gong}{s.gong@qmul.ac.uk}{1}

\addinstitution{
Computer Vision Group,\\
School of Electronic Engineering and Computer Science,\\
Queen Mary University of London,\\
London, E1 4NS, UK
}

\runninghead{Jiabo Huang and Shaogang Gong}{Semantic Contrastive Learning}

\usepackage{epsfig}
\usepackage{graphicx}
\usepackage{amsmath}
\usepackage{amssymb}
\usepackage{bm}
\usepackage{xspace}
\usepackage{xcolor}
\usepackage{bbm}
\usepackage{algorithm}
\usepackage{pifont}
\usepackage{varwidth}
\usepackage{capt-of}
\usepackage{multirow}
\usepackage{wrapfig}
\graphicspath{{./figures/}}

\renewcommand{\paragraph}[1]{\vspace{0.1cm}\noindent\textbf{#1}\quad}

\def\fullname{Semantic Contrastive Learning\xspace}
\def\abbrname{SCL\xspace}

\def\eg{\emph{e.g}\bmvaOneDot} 
\def\ie{\emph{i.e}\bmvaOneDot} 
 
 \def\vs{\emph{vs}\bmvaOneDot}
\def\wrt{w.r.t\bmvaOneDot}

\DeclareMathOperator{\softmax}{\text{Softmax}}

\newcommand{\best}[1]{{\textbf{\color{red}#1}}}
\newcommand{\scnd}[1]{{\underline{\color{blue}#1}}}

\begin{document}

\maketitle

\begin{abstract}
Whilst contrastive learning has recently brought notable benefits to
deep clustering of unlabelled images 
by learning sample-specific discriminative visual features, 
its potential for
explicitly inferring class decision boundaries is less well understood.
This is because its instance discrimination strategy is not class sensitive, 
hence,
the clusters derived on the resulting feature space
are not optimised for
corresponding to meaningful class decision boundaries. 
In this work, 
we solve this problem by introducing Semantic Contrastive Learning (SCL). 
SCL imposes explicitly distance-based
cluster structures on unlabelled training data 
by formulating a semantic (cluster-aware) contrastive learning objective.
Specifically, we encourage consensus between learning the optimal
hypotheses on the semantic class boundaries and feature similarities. 
This is formulated by a clustering consistency condition
to be satisfied jointly by \textit{instance} feature similarities and
\textit{cluster} decision boundaries.
This semantic contrastive learning approach 
to discovering unknown class decision boundaries 
has considerable advantages to unsupervised learning of object recognition. 
Extensive experiments show 
that SCL outperforms state-of-the-art 
contrastive learning and deep clustering methods 
on six object recognition benchmarks,
especially on the more challenging finer-grained and larger datasets.
\end{abstract}

\section{Introduction}
\label{sec:intro}
Given the massive increase of images available on the Internet,
how to leverage them without label annotation
for learning high-level visual semantics 
remains a challenging problem for unsupervised deep learning, 
although it has been shown to be highly effective in 
supervised deep learning given large-scale labelled training data.
Clustering as a conventional unsupervised machine learning technique%
~\cite{achanta2017superpixels,joulin2010discriminative,liu2018learning} 
has been recently exploited for visual representation learning 
in deep neural networks to perform \textit{Deep Clustering}%
~\cite{hershey2016deep,huang2020pica}.

\begin{figure}[ht]
\begin{center}
\includegraphics[width=1.0\linewidth]{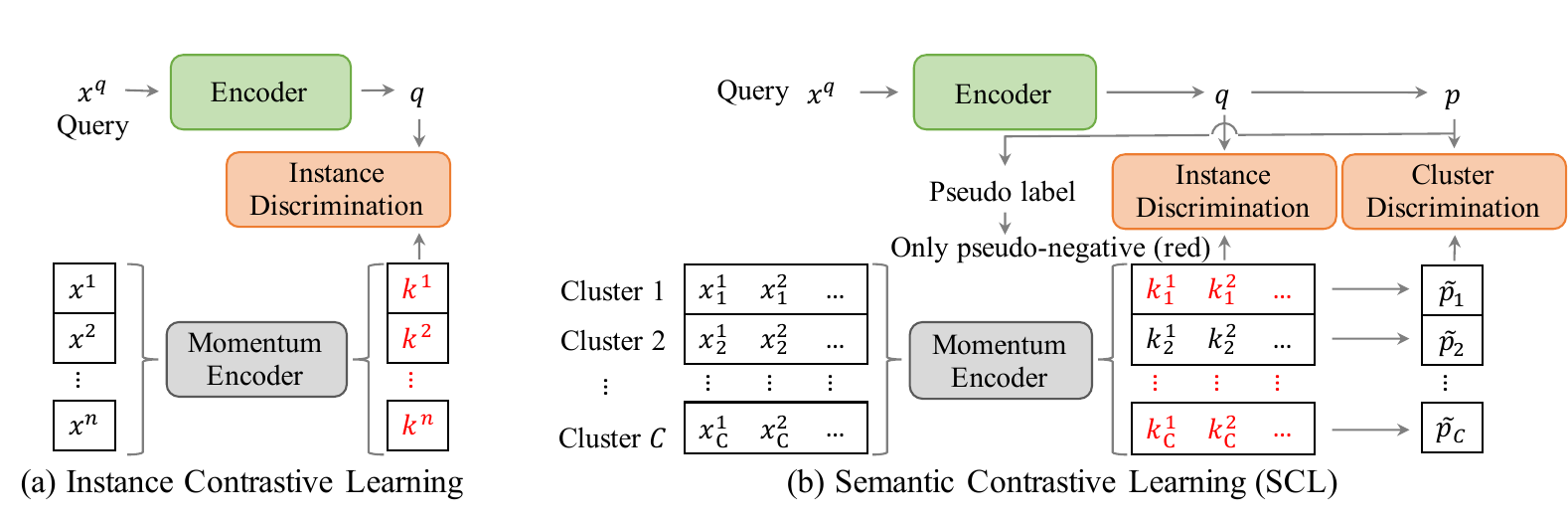}
\end{center}
\caption{
Instance \vs semantic contrastive learning.
(a) Instance contrastive learning differentiates query samples 
from a contrastive set regardless their potential class memberships.
(b) SCL pulls samples away from only their pseudo negatives 
of other clusters.
}
\label{fig:idea}
\end{figure}

Separately,
contrastive learning~\cite{he2020moco,chen2020simclr,tian2019CMC}
has also been shown effective for self-supervised learning of 
\textit{generalisable} feature representations
by \textit{instance-discrimination} (Fig.~\ref{fig:idea} (a)). 
It may appear to have the potential to benefit 
unsupervised clustering due to 
its expressive sample-specific visual representation. 
However, 
directly applying contrastive learning to deep clustering 
is sub-optimal for \textit{class-discrimination}. 
Because it lacks awareness of nonlinear intra-class variations.
This is inherent
from learning with per-sample pseudo classes 
generated by global linear data augmentations
for instance-wise visual discrimination.
We observe that this limitation 
is overlooked by recent contrastive learning based clustering methods. 
Such attempts avoid the problem by 
either limiting to restricted local feature space neighbourhoods 
exhibiting subtle visual variations~\cite{huang2018and,huang2019pad},
or suffering from class ambiguities 
due to the inherent contradiction 
between instance-level discrimination 
(pull away intra-cluster) 
and class(cluster)-level grouping 
(push closer intra-cluster)%
~\cite{li2021contrastive,van2020scan,tao2021clustering}.

In this work,
we propose a deep clustering method called \textit{\fullname} (\abbrname).
In this \abbrname model, 
cluster structures are explicitly imposed to unlabelled training data
to encourage learning a `cluster-aware' instance discriminative feature space
that promotes separation of decision boundaries between clusters, 
leading to a plausible interpretation of
the underlying semantic concepts (Fig.~\ref{fig:idea} (b)).
Specifically, 
the instance discrimination in SCL 
aims to reduce visual redundancy (what's common) \textit{across samples} 
so that images which are sharing more uncommon (what's unique) appearance patterns
are pushed closer in feature space,
whilst the cluster discrimination 
aims to optimise holistically the cluster decision boundaries 
so that any visual overlap \textit{across clusters} 
is minimised and each cluster exhibits unique and
consistent visual characteristics 
of each underlying class (semantic concept).
Different from
the recent instance contrastive learning based clustering methods~\cite{van2020scan,li2021contrastive,tao2021clustering}
which pull away each instance from \textit{all} other samples 
in the feature space,
\abbrname only pulls it away from its \textit{pseudo-negative} samples in other clusters.
By sharing a common contrastive (negative) set for all the instances in a cluster,
\abbrname indirectly pushes them closer regardless of any intra-cluster visual dissimilarity.
This resolves the contradiction in the
instance contrastive learning and clustering objectives
but is neglected by those recent attempts.
Moreover,
we introduce a new semantic memory
to not only store representations for instance discrimination
but also embed the cluster structures. 
This enables optimising cluster decision boundaries by 
maximising the consistency between
cluster-level (semantic) and instance-level (visual) distances. 

Our \textbf{contributions} are:
\textbf{(1)}
We make the first attempt to
solve the contradiction in learning simultaneously instance
contrastive discrimination and clustering objectives in order to
optimise nontrivial class separations in a feature space without 
labelled training.
\textbf{(2)}
We introduce a novel \textit{\fullname} (\abbrname) for 
deep clustering. 
\abbrname discovers cluster decision boundaries 
by enforcing a consensus
between instance contrastive discrimination 
and cluster compactness. 
\textbf{(3)}
We formulate a new semantic memory
to enable simultaneous optimisation of 
instance and cluster discrimination. 
\abbrname yields compelling performance advantages
over the state-of-the-art deep clustering methods,
with significant improvements ($\sim$17\%)
on the more challenging larger and finer-grained datasets.

\section{Related Work}
\label{sec:review}
We shall first differentiate clearly 
the different objectives between deep clustering in the context of this work and
unsupervised representation learning elsewhere. 
The latter aims to learn 
generalisable feature representations 
-- generative representational learning -- 
without any consideration for optimising class-discrimination. 
Our objective is generative decision boundary learning 
optimised for class-discrimination without labels in model learning.

\paragraph{Deep clustering.}
In the absence of ground-truth class labels,
one popular solution of deep clustering is
to mimic supervised learning by estimating pseudo labels iteratively
from learning improvement on feature representations%
~\cite{xie2016dec,guo2017idec,yang2016jule,yang2017towards,chang2017dac,chang2019ddc,wu2019dccm}.
Although these methods may benefit from explicit supervised discriminative learning, 
it is also intrinsically unstable due to
error-propagation between unreliable label assignments and updates of
randomly initialised representation 
based on such assignments~\cite{huang2018and,zou2019confidence}.
\abbrname is more robust to error propagation
from the intermediate cluster assignments during model learning
because the contrastive learning formulation
is able to discover the intrinsic visual similarity among samples
despite a lack of knowledge of their true class memberships.
In contrast to the alternate strategy,
one can learn simultaneously label assignment
and feature updates using certain pretext objectives that indirectly
impose requirements for learning good cluster structures%
~\cite{ji2017subspace,peng2017subspace,haeusser2018adc,xu2019iic,huang2020pica,zhao2020dccs,niu2020gatcluster}.
However, due to the weak correlations between their learning objectives
and the target class boundary separations,
they tend to yield clusters that are less consistent with the
semantic categories.
\abbrname reduces visual redundancy across clusters 
so that each cluster exhibits unique and
consistent visual characteristics that are more plausible for encoding 
an underlying semantic concept.

There are a few recent attempts%
~\cite{van2020scan,li2021contrastive,tao2021clustering,do2021clustering}
on deep clustering by exploring directly visual features from
instance contrastive learning.
However,
they either suffer from class ambiguities due to the inherent
contradiction between 
instance-level discrimination (pull away intra-cluster) and 
cluster-level grouping (push closer intra-cluster)%
~\cite{li2021contrastive,do2021clustering},
or focused only on a one-sided representation learning~\cite{tao2021clustering} 
or their partitioning~\cite{van2020scan} 
while neglecting their mutual impacts.
By assembling instance-wise contrastive samples 
into a common pseudo-negative set 
for simultaneous instance discrimination 
and cluster decision boundary optimisation, 
we resolve their contradiction and jointly amplify their strengths.

\paragraph{Unsupervised representation learning.}
Beyond the forementioned works designed for
modelling the inherent class structure of unlabelled images,
there are other methods for learning generalisable image representations
that may appear to be similar to clustering%
~\cite{Caron2018deepclustering,caron2019unsupervised,caron2020unsupervised,asano2019self,li2021PCL}.
Those representation learning methods assume clustering \textit{is given},
which rely on independent clustering%
~\cite{Caron2018deepclustering,caron2019unsupervised,li2021PCL} or 
optimal transport algorithms~\cite{asano2019self,caron2020unsupervised}
to compute the pseudo labels. 
Therefore, they are both limited by 
potentially suboptimal clustering computed independently,
and only addressing restricted partial problem, an easier learning task.
Our \abbrname model solves the two underlying problems
holistically as a single problem by focusing directly on modelling
semantically-aware clustering therefore removing any suboptimal
offline clustering assumption and is end-to-end optimised for the
resulting representation derived. 

Contrastive learning%
~\cite{wu2018instance,he2020moco,chen2020simclr,chen2020moco_v2,tian2019CMC,oord2018representation} 
optimises sample-specific visual features
by treating every individual instance as an independent class augmented by
guaranteed positive samples generated using global linear transforms.
By ignoring any cross-sample relationships and global class memberships,
the learned representations are ambiguous to both 
intra and inter-class nonlinear image variations,
therefore, less discriminative against true classes.
To address such a limitation,
studies have been carried out to integrate it with 
neighbourhood discovery~\cite{huang2018and,huang2019pad,zhuang2019local}.
These methods 
adopt directly the supervised contrastive learing~\cite{khosla2020supervised} paradigm
to \textit{explicitly} push pseudo-positive samples \textit{closer}
in the feature space.
Such a paradigm is prone to accumulating errors 
from unreliable pseudo label predictions.
Extra constraints and strategies 
such as restricting neighbourhood's size
and pre-learning representations
must be applied
to avoid the negative impacts of error-propagation. 
Such strategies cannot apply in general therefore are suboptimal.
In contrast, 
our \abbrname model \textit{implicitly} poses
positive relationships by pulling samples \textit{away} from a common
pseudo-negative contrastive set. 
\abbrname has no need
for hand-crafted extra strategies 
which are time consuming and non-scalable
due to being independent from the deep clustering learning, not end-to-end.

%
\begin{figure*}[ht]
\begin{center}
\includegraphics[width=1.0\linewidth]{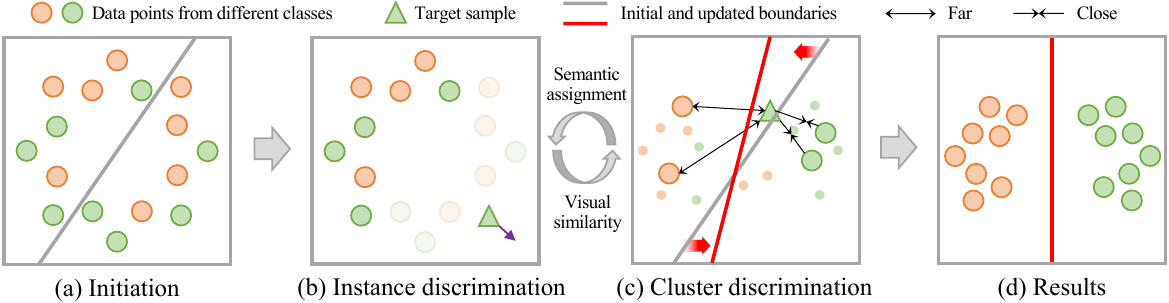}
\end{center}
\caption{An overview of \abbrname.
\textbf{(a)}
Given a randomly initialised feature space and decision boundaries,
\textbf{(b)}
the \abbrname model optimises visual similarities among samples
by instance discrimination and
\textbf{(c)}
the potential class memberships by cluster discrimination
\textbf{(d)}
and finally converge to a consensus
between instance-level diversity and cluster-level compactness.
}
\label{fig:pipeline}
\end{figure*}
\section{Clustering on Unlabelled Images}
\label{sec:method}
Given a set of \textit{unlabelled} images 
$\mathcal{I} = \{\bm{I}_1, \bm{I}_2, \cdots, \bm{I}_N\}$,
deep clustering aims to
derive (1) a {\em feature embedding network} $\theta$
that extracts key semantic information 
encoded in the high-dimensional pixel space to a compact vector subspace
$f_\theta: \bm{I} \rightarrow \bm{x} \in \mathbb{R}^d$,
and (2) a \textit{classifier} $\phi$
that projects the feature vectors into $C$ partitions
$f_\phi: \bm{x} \rightarrow y, y \in \{1, 2, \cdots C\}$,
with a hope that samples in the same cluster share the same ground-truth class label,
otherwise not.
It is fundamentally challenging 
to derive class discriminative information directly
from raw images in an unsupervised manner,
due to the complex appearance patterns and variations exhibited
both within and across classes.

In this work,
we introduce a \textit{\fullname} (\abbrname) method.
\abbrname explores the idea of cluster-centred contrastive learning
that differ from other recent developments on deep clustering by
directly applying instance-centred contrastive learning.
Given the sample-specific learning constraint 
of the instance contrastive learning, 
it is non-trivial to exploit it in unsupervised clustering 
that also jointly enforces necessary constraints 
to unknown class decision boundary 
when there is no class label in training.
To overcome this hurdle, 
we 
optimises concurrently
instance discrimination and their assignment to a set of clusters, 
with an additional consistency 
objective function 
to condition their optimisations jointly. 
\abbrname aims to learn both optimal instance visual similarities
that can verify each instance's cluster assignment globally and 
optimal cluster compactness that can maximise inter-cluster discrimination
margins.
Importantly, the \abbrname formulation can be utilised by
any instance contrastive learning methods~\cite{wu2018instance,he2020moco,chen2020simclr}
for deep clustering tasks, and it is end-to-end trainable therefore
globally optimised.
Fig.~\ref{fig:pipeline} shows an overview of \abbrname.


\subsection{Semantic Contrastive Learning}

We start with formulating a new
\textit{cross-cluster} instance discrimination learning objective
with a novel \textit{semantic memory}.
The aim is to learn visual features
to be discriminative across clusters
and
facilitate simultaneous instance and cluster discrimination.

\paragraph{Cross-cluster instance discrimination.}
Our feature learning objective is formulated
to differentiate every individual instance against its pseudo-negative samples
so to reduce its visual redundancy 
regarding images of other clusters
(Fig.~\ref{fig:pipeline} (b)).
Given random partitions at the beginning of training
(Fig.~\ref{fig:pipeline} (a)),
by isolating samples from different clusters,
the model behaves as instance contrastive learning
and outputs per sample-specific visual features.
Intuitively,
visually similar samples are expected to share more
class-specific unique information,
their representations will therefore be gradually gathered closer
and grouped into the same clusters by our cluster discrimination detailed later.
Along the clustering process with increasingly better and stable cluster assignments,
the contrastive set of every sample
will absorb more visually dissimilar counterparts,
instead of random ones.
Consequently,
the learning objective becomes 
reducing cross-cluster visual redundancy,
resulting in desired features that are 
aware of inter-cluster visual discrepancies
and invariant within clusters (Fig.~\ref{fig:pipeline} (d)).


Whilst our \abbrname is a generic formulation,
we take the momentum contrast (MoCo)~\cite{he2020moco,chen2020moco_v2} 
as an example of instantiation.
We first formulate a mapping function $f_\theta$ 
from a pixel space to a representational space
as an encoder with learnable weights $\theta$.
Similarly, we construct another momentum encoder $f_{\tilde{\theta}}$ 
with an identical structure but independent parameters $\tilde{\theta}$.
Given an unlabelled dataset $\mathcal{I}$,
we randomly apply a set of transformations $\mathcal{T}$ to each image for distribution perturbation.
We then represent two perturbed copies of each instance,
$\mathcal{T}_1(\bm{I}_i)$ and $\mathcal{T}_2(\bm{I}_i)$,
by the two encoders respectively
and denote them as 
$\bm{q}_i = f_\theta(\mathcal{T}_1(\bm{I}_i))$ 
and $\bm{k}_i = f_{\tilde{\theta}}(\mathcal{T}_2(\bm{I}_i))$.
Given the pseudo labels of all the samples
$\mathcal{Y} = \{y_1, y_2, \cdots, y_N\},\ y_i \in [1, C]$
inferred by
the progressively updating decision boundaries, 
our instance discrimination objective in terms of $\bm{I}_i$
is to match $\bm{q}_i$ with $\bm{k}_i$
against its contrastive set 
${Q}_i = \{\tilde{\bm{k}}_1, \tilde{\bm{k}}_2, \cdots, \tilde{\bm{k}}_K\}\ s.t.\ y_i \neq y_j, \forall j \in [1,K]$
composed by $K$ stale representations of its pseudo-negative samples:
\begin{equation}
\mathcal{L}_\text{ID}(\bm{I}_i) = -\log
\frac{\text{exp}(cos(\bm{q}_i, \bm{k}_i) / \tau)}{\sum_{\tilde{\bm{k}} \in Q_i \cup \{\bm{k}_i\}}\text{exp}(cos(\bm{q}_i, \tilde{\bm{k}}) / \tau)},
\label{eq:mi_loss}
\end{equation}
where $cos(\cdot, \cdot)$ is the cosine similarity between a pair of representations
and $\tau$ is the temperature to control the concentration degree of distribution.
As the samples in the same clusters
share a common contrastive set,
they are indirectly pushed closer in the feature space
regardless of any intra-cluster variations.
Therefore,
the learned features are geared towards being
sensitive to cluster-wise visual characteristics, 
not sample-wise.

\paragraph{Semantic memory.}
To facilitate instance discrimination across clusters,
we manage $C$ independent memory banks $\mathcal{M} = \{{M}_1, {M}_2, \cdots {M}_C\}$ each 
corresponding to one cluster with a size of $K / (C - 1)$. 
For an image $\bm{I}_i$ with pseudo label $y_i$,
we construct its contrastive set ${Q}_i$:
\begin{equation}
{Q}_i = \{\tilde{\bm{k}} | \tilde{\bm{k}} \in {M}_j\ \forall j \in [1,C]\ \text{and}\ j \neq y_i\}.
\label{eq:memory}
\end{equation}
There is always one memory bank left out for each sample
and the rest ${M}$s are concatenated as its contrastive set ${Q}$ approximately in size $K$ (rounding error)
to support cluster discriminative feature representation learning.
For memory update, 
after every backward pass,
the representation $\bm{k}_i$ enqueues
to ${M}_{y_i}$
with the oldest one inside removed.

\paragraph{Cluster discrimination.}
To discover the underlying concepts with unique visual characteristics,
we infer their decision boundaries by
reducing the visual redundancy among clusters,
namely maximising the visual similarity of samples within the same clusters
and minimising that between clusters
(Fig.~\ref{fig:pipeline} (c)).
Concretely,
as the representation of samples with different pseudo labels
are stored independently in the semantic memory bank,
they can be taken as anchors to describe their corresponding clusters.
Given a training sample $\bm{q}_i$,
its probability $\tilde{p}_{i,j}$ of being in the $j$-the cluster
predicted by a distance-based classifier is
\begin{equation}
\tilde{p}_{i,j} = \frac{\sum_{\tilde{\bm{k}} \in M_j}\text{exp}(cos(\bm{q}_i, \tilde{\bm{k}}) / \tau)}{\sum_{j'=1}^C\sum_{\tilde{\bm{k}} \in M_{j'}}\text{exp}(cos(\bm{q}_i, \tilde{\bm{k}}) / \tau)}.
\label{eq:semantic_memory_prediction}
\end{equation}
With such potential memberships determined by
sample-anchor visual similarities,
we formulate a consistency loss 
for learning the cluster decision boundaries:
\begin{equation}
\bm{p}_i = \softmax({W}^\top \bm{q}_i + {B})\  
\in \mathcal{R}^C,\quad
\mathcal{L}_\text{CD} = \frac{1}{n}\sum_{i=1}^n\sum_{j=1}^C -\tilde{p}_{i,j}\log p_{i,j},
\label{eq:proto_loss}
\end{equation}
where $\{W;B\}$ is the learnable parameters of classifier $f_\phi$
and $n$ denotes the size of mini-batch.
In Eq.~\eqref{eq:proto_loss},
we aim to minimise the cross-entropy of
the distance-based cluster assignments $\tilde{\bm{p}}_i$
and the predictions $\bm{p}_i$ yielded by the cluster decision boundaries
then propagate the gradient back to $\bm{p}_i$ only
to avoid feature learning from unreliable boundaries.
By doing so,
samples are assigned to the cluster with the most similar anchors
while each cluster holding its own visual characteristics
that make it different from others
and correspond to an underlying semantic class
with consistent and unique visual characteristics.

With the updated models $f_\theta$ and $f_\phi$,
we renew the cluster assignments every epoch
in a maximum likelihood manner
for semantic memory construction in Eq.~\eqref{eq:memory}:
\begin{equation}
y_i = \underset{j}{\arg\max}\ p_{i,j},\ j\ \in\ \{1, 2, \cdots, C\}.
\label{eq:label}
\end{equation}
As the predictions become increasingly more accurate 
in the process of training, 
this update improves cross-cluster instance
discrimination on learning class discriminative features.

\paragraph{Hard samples mining.}
To enhance
discrimination capacity,
we identify semantically ambiguous samples
and emphasise them in instance discrimination:
\begin{gather}
s_i^e = s_i^{e-1} + \mathbbm{1}[y_i^{e} \neq y_i^{e-1}],\quad
w_i^e = \frac{s_i^e}{\sum_j^n s_j^e},\quad
\mathcal{L}_\text{ID} = \sum_{i=1}^n w_i^e \mathcal{L}_\text{ID}(\bm{I}_i), \label{eq:reweight_loss}
\end{gather}
where $w_i^e$ is the weights of $\bm{I}_i$ at the $e$-th training epoch.
The samples that are frequently swapped across clusters (\ie hard samples) are assigned with higher weights for offering more useful discriminative learning clues.

\subsection{Model Training}

Given the instance (Eq.~\eqref{eq:reweight_loss}) and cluster (Eq.~\eqref{eq:proto_loss}) discrimination losses,
the overall training objective of \abbrname is:
\begin{equation}
\mathcal{L} = \alpha\mathcal{L}_\text{ID} + \beta\mathcal{L}_\text{CD}.
\label{eq:overall_loss}
\end{equation}
In the absence of labelled validation data in unsupervised clustering,
we set both the weights to $\alpha\!=\!\beta\!=\!1$ to avoid exhaustive per-dataset parameter tunning.
To minimise $\mathcal{L}$,
the weights of encoder $\theta$ as well as the decision boundaries $\phi$ are updated by back-propagation
and the momentum encoder $\tilde{\theta}$ is by 
$\tilde{\theta} \leftarrow m\tilde{\theta} + (1-m)\theta$
where $m$ is a momentum coefficient~\cite{he2020moco}.
Both objective functions (Eq.~\eqref{eq:reweight_loss} and Eq.~\eqref{eq:proto_loss}) are differentiable
thus can be trained end-to-end by the conventional stochastic gradient descent algorithm.

\section{Experiments}
\label{sec:exp}

\paragraph{Datasets.}
Evaluations were conducted on six challenging object recognition benchmarks.
\textbf{(1) CIFAR-10(/100)~\cite{krizhevsky2009cifar}}:
Natural image datasets composed by 60,000 samples 
that are uniformly drawn from 10(/100) classes. The 20 super-classes on CIFAR-100 were considered as ground-truth.
\textbf{(2) STL-10~\cite{coates2011stl}}:
An ImageNet adapted dataset consists of 1,300 images
from each of 10 classes.
Additional 100,000 images from unknown classes
were available but deprecated in our experiments.
\textbf{(3) ImageNet-10/Dogs~\cite{russakovsky2015imagenet}}:
ImageNet subsets containing samples from 
10 randomly selected classes or 15 dog breeds.
\textbf{(4) Tiny-ImageNet~\cite{le2015tiny_imagenet}}:
Another ImageNet subset in larger-scale with 100,000 samples 
evenly distributed in 200 classes.
Training and testing are conducted on the same set of data following convention~\cite{xu2019iic,huang2020pica}.


\paragraph{Evaluation metrics.}
Three standard clustering metrics
were used to measure the consistency of cluster assignments
and ground-truth class memberships:
(1) Clustering accuracy (\textbf{ACC})
maps one-to-one the learned clusters
to the ground-truth classes by the Hungarian algorithm~\cite{kuhn1955hungarian}
and measures the classification accuracy;
(2) Normalised mutual information (\textbf{NMI})
quantifies the labelling consistency by the normalised MI
between the predicted and ground-truth labels of all image samples;
(3) Adjusted rand index (\textbf{ARI}) computes the ratio of
image sample pairs
holding consistent pairwise relationships against the ground-truth.
All these metrics scale from $0$ to $1$
and higher is better.

\paragraph{Implementation details.}
We followed~\cite{xu2019iic,huang2020pica}
to use a variant of ResNet-34 as the backbone network
and~\cite{chen2020moco_v2} for the other implementation choices.
All our models and the cluster assignments are randomly initialised.
An SGD optimiser was adopted for model updates with weight decay in $5e-4$.
The coefficient for momentum encoder updating was $0.9$ and 
$\tau$ in Eq.~\eqref{eq:mi_loss} was $0.1$.
We stored $4096 / (C - 1)$ representations for each cluster 
in the semantic memory (Eq.~\eqref{eq:memory})
on all the datasets except for $8192 / (C - 1)$ on Tiny-ImageNet due to larger scale. 
The learning rate was set to $0.03$ 
with the cosine schedule~\cite{loshchilov2016sgdr} 
for its adjustment across $200$ epochs
while the batch size was $256$.
We adopted the ``merge-and-split'' strategy~\cite{zhan2020online}
for updating pseudo labels (Eq.~\eqref{eq:label})
to avoid extremely imbalanced partitions
and to stabilise training.
Besides the target `clustering' tasks which partition 
the target data into the ground-truth number of clusters
to facilitate comparisons, 
we followed~\cite{xu2019iic,huang2020pica}
to jointly train \abbrname with auxiliary `under-clustering' and `over-clustering' tasks
so to explore multi-grained visual similarity.
The cluster number in `under-clustering' was half of the ground-truth 
while instance-wise learning was considered as extreme `over-clustering'.
At test time, 
we followed~\cite{tao2021clustering,li2021contrastive} 
to compare by the best models
using the assignments
yielded by the classifier for `clustering' tasks
while the other two were deprecated.
All the hyper-parameters were kept the same across different datasets,
\ie no exhaustive per dataset tuning.
On computational cost, 
the only extra parameters we introduced to MoCo~\cite{chen2020moco_v2}
are in the linear classifier $f_\phi$
and it took around 30 seconds on CIFAR-10 to 
update pseudo labels per epoch.

\begin{table}[t] 
\scriptsize
\setlength{\tabcolsep}{2.3pt}
\centering
\begin{tabular}{l|ccc|ccc|ccc|ccc|ccc|ccc}
\hline
Dataset
& \multicolumn{3}{c|}{CIFAR-10}
& \multicolumn{3}{c|}{CIFAR-100}
& \multicolumn{3}{c|}{STL-10}
& \multicolumn{3}{c|}{ImageNet-10}
& \multicolumn{3}{c|}{ImageNet-Dogs}
& \multicolumn{3}{c}{Tiny-ImageNet} \\
\hline  
Metrics
& NMI & ACC & ARI & NMI & ACC & ARI
& NMI & ACC & ARI & NMI & ACC & ARI
& NMI & ACC & ARI & NMI & ACC & ARI \\ \hline\hline
K-means
& .087 & .229 & .049 & .084 & .130 & .028 
& .125 & .192 & .061 & .119 & .241 & .057
& .055 & .105 & .020 & .065 & .025 & .005 \\ \hline
DEC$^*$~\cite{xie2016dec}
& .257 & .301 & .161 & .136 & .185 & .050 
& .276 & .359 & .186 & .282 & .381 & .203  
& .122 & .195 & .079 & .115 & .037 & .007 \\ \hline
DAC$^*$~\cite{chang2017dac}
& .396 & .522 & .306 & .185 & .238 & .088 
& .366 & .470 & .257 & .394 & .527 & .302  
& .219 & .275 & .111 & .190 & .066 & .017 \\ \hline
ADC$^*$~\cite{haeusser2018adc} 
&   -   & .325 &   -   &   -   & .160 &   - 
&   -   & .530 &   -   &   -   &   -   &   -
&   -   &   -   &   -   &   -   &   -   &   -   \\ \hline
DDC$^*$~\cite{chang2019ddc}
& .424 & .524 & .329 &   -   &   -   &   -
& .371 & .489 & .267 & .433 & .577 & .345
&   -   &   -   &   -   &   -   &   -   &   -   \\ \hline
DCCM$^*$~\cite{wu2019dccm}
& .496 & .623 & .408 & .285 & {.327} & {.173} 
& .376 & .482 & .262 & .608 & .710 & .555 
& .321 & {.383} & {.182} & .224 & {.108} & {.038} \\ \hline
IIC~\cite{xu2019iic} 
& .513 & .617 & .411 &   -   & .257 &   - 
& .431 & .499 & .295 &   -   &   -   &   -
&   -   &   -   &   -   &   -   &   -   &   -   \\ \hline
PICA~\cite{huang2020pica}
& .591 & .696 & .512 & {.310} & .337 & .171
& .611 & .713 & .531 & {.802} & {.870} & {.761}
& {.352} & .352 & .201 & {.277} & .098 & .040 \\ \hline
DCCS$^*$~\cite{zhao2020dccs}
& {.569} & {.656} & {.469} &   -   &   -   &   -
& .376 & .482 & .262 & .608 & .710 & .555
&   -   &   -   &   -   &   -   &   -   &   - \\ \hline
GAT$^*$~\cite{niu2020gatcluster}
& .475 & .610 & .402 & .215 & .281 & .116
& {.446} & {.583} & {.363} & .594 & .739 & .552
& .281 & .322 & .163 &   -   &   -   &   - \\ \hline\hline
SCAN$^\dagger$~\cite{van2020scan}
& {.712} & \scnd{.818} & {.665} & {.441} & .422 & {.267}
& .654 & .755 & .590 &   -   &   -   &   -  
&   -   &   -   &   -   &   -   &   -   &   -   \\ \hline
IDFD$^\dagger$~\cite{tao2021clustering}
& .711 & {.815} & .663 & .426 & .425 & .264
& .643 & .756 & .575 & \best{.898} & \best{.954} & \best{.901}
& \scnd{.546} & \scnd{.591} & \scnd{.413} &   -   &   -   &   -   \\ \hline
CC$^\dagger$~\cite{li2021contrastive}
& .705 & .790 & .637 & {.431} & {.429} & {.266}
& \best{.764} & \best{.850} & \best{.726} & .859 & .893 & .822
& .445 & .429 & .274 & {.340} & \scnd{.140} & {.071} \\ \hline
CRLC$^\dagger$~\cite{do2021clustering}
& .679 & .799 & .634 & .416 & .425 & .263
& \scnd{.729} & \scnd{.818} & \scnd{.682} & .831 & .854 & .759
& .461 & .484 & .297 &   -   &   -   &   -   \\ \hline
GCC$^\dagger$~\cite{do2021clustering}
& \best{.764} & \best{.856} & \best{.728} & \scnd{.472} & \scnd{.472} & \scnd{.305}
& .684 & .788 & .631 & .842 & .901 & .822
& .490 & .526 & .362 & \scnd{.347} & .138 & \scnd{.075}  \\ \hline
\textbf{\abbrname}$^\dagger$$^*$
& \scnd{.744} & {.813} & \scnd{.683}
& \best{.477} & \best{.482} & \best{.314}
& {.593} & {.638} & {.485}
& \scnd{.877} & \scnd{.930} & \scnd{.861}
& \best{.728} & \best{.763} & \best{.652}
& \best{.337} & \best{.172} & \best{.080} \\
\hline
\end{tabular}
\caption{
Comparisons to the state-of-the-art deep clustering approaches.
Methods with $(\cdot)^\dagger$
conducted deep clustering by contrastive learning
and $(\cdot)^*$ trained without the additional data on STL-10.
The $1^\text{st}$/$2^\text{nd}$ best results
are highlighted in \best{red}/\scnd{blue}.
}
\label{tab:sota_clustering}
\end{table}

\subsection{Comparisons to the State-of-the-Art}
\paragraph{Deep Clustering.}
Table~\ref{tab:sota_clustering} 
compares the proposed \abbrname
with a wide range of state-of-the-art deep clustering models 
including both with- (from ``SCAN'' and below) and 
without- (from ``GAT''and above) contrastive learning in their formulation.
We observe:
\textbf{(1)}
\abbrname has broad advantages
over all other methods including the few close competitors, 
\eg by $17.2\%$ (ACC) improvement over IDFD on ImageNet-Dogs. 
\abbrname yielded the best results in 4 out of the 6
benchmarks and at least top-2 in 5/6 benchmarks.
On STL-10 where SCL seems less competitive, 
the top models used almost 10 times more additional training images
that are sampled from the same distribution as the target data
but are explicitly of different classes independent from the target classes. 
Those additional data give significant benefits by learning from
strong negative signals but then were explicitly excluded when
training the target classifier, making it an easier learning task.
In our experiment,
we avoided using such a data engineering
strategy because it is neither practical nor scalable to have 
such similar and guaranteed negative data 
unless their class labels are available. 
On the other hand,
\abbrname's performance advantages over those methods 
learned without the additional data engineering
(marked with $*$) remain notable, 
\eg improving GAT by $5.5\%$.
This is a more accurate reflection on models' true
performances which is also consistent to the other benchmarks.
\textbf{(2)}
It is always more challenging to precisely model the truth
class boundaries of either 
finer-grained or larger datasets.
In these cases, \abbrname surpassed 
IDFD and CC
on ImageNet-Dogs and Tiny-ImageNet
by $17.2\%$ and $3.2\%$, respectively.
\textbf{(3)}
The significant performance margins obtained by all
contrastive learning based methods
indicate compellingly the benefit of contrastive constraints in unsupervised
semantic concepts learning.
Importantly,
SCL's superiority 
demonstrate the significance of solving the contradiction
between optimising instance contrastive discrimination (pull apart) 
and intra-cluster compactness (push closer). 

\begin{table}[t]\footnotesize
\begin{minipage}[b]{0.48\linewidth}
\centering
\setlength{\tabcolsep}{2pt}
\begin{tabular}[b]{l|c|c|c}
\hline
Dataset & CIFAR-10 & CIFAR-100 & STL-10 \\ \hline
\hline\hline
MoCo~\cite{chen2020moco_v2}$^\star$ & 0.528 & 0.360 & 0.561 \\ \hline
PAD~\cite{huang2019pad}$^\dag$ & 0.626 & 0.288 & 0.465 \\ \hline
DeepCluster~\cite{Caron2018deepclustering}$^\dagger$ & 0.374 & 0.189 & 0.334 \\ \hline
\textbf{\abbrname} & \best{0.813} & \best{0.482} & \best{0.638} \\
\hline
\end{tabular}
\caption{
Comparisons to representation learning methods. 
Notation: $(\cdot)^\star$ indicates results reproduced 
from scratch using the authors' code~\cite{chen2020moco_v2};
$(\cdot)^\dag$ are from~\cite{huang2019pad}.
}
\label{tab:sota_contrasting}
\end{minipage}%
\hfill
\begin{minipage}[b]{0.48\linewidth}
\centering
\includegraphics[width=\textwidth]{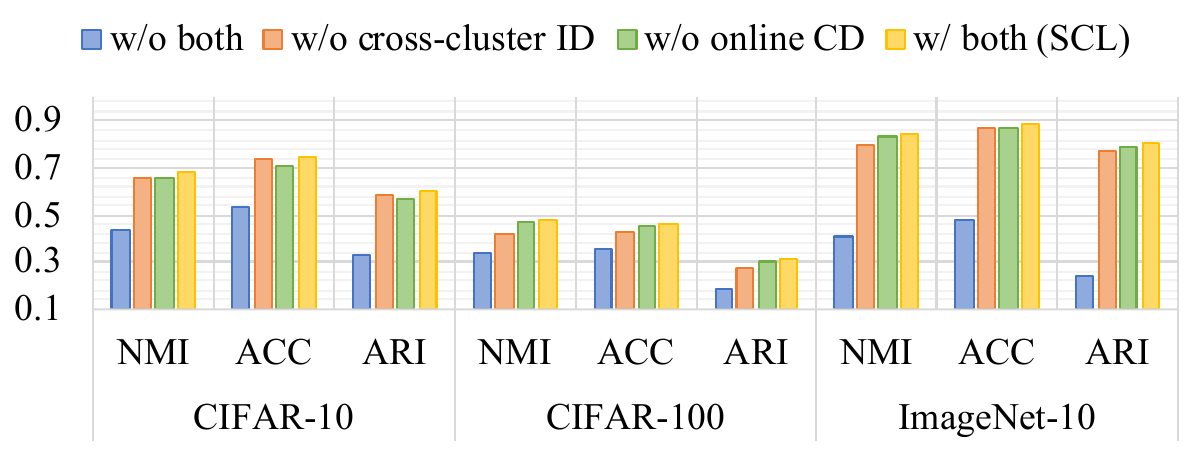}
\captionof{figure}{Ablation studies on \textit{cross-cluster} instance descrimination (ID) and 
\textit{online} cluster discrimination (CD) designs.
}
\label{fig:ablation}
\end{minipage}
\end{table}

\paragraph{Representation learning.}
Beyond the methods intrinsically designed for clustering%
~\cite{van2020scan,tao2021clustering,li2021contrastive,do2021clustering},
we also compared \abbrname with
a clustering-based representation learning approach~\cite{Caron2018deepclustering}
and
two general instance contrastive learning schemes:
Instance-wise learning (MoCo~\cite{chen2020moco_v2}) and
local neighbourhood discrimination based learning (PAD~\cite{huang2019pad}).
The learned feature representations from both models are applied with K-means for clustering.
As shown in Table~\ref{tab:sota_contrasting},
our \abbrname method outperformed all the representation learning methods across the board.
This shows clearly 
the advantages of \abbrname from
holistically modelling the inherent class structure,
resulting also a more optimal representation, 
as compared to separating representation learning from class membership estimation.

\subsection{Ablation Study}
Detailed ablation studies were conducted
for in-depth analysis of \abbrname.
K-means was adopted for models
which did not yield desired number of clusters.
Experimental results were averaged over multiple trials.

\paragraph{Instance and cluster discrimination.}
We investigated the independent contributions
of our \textit{cross-cluster} instance discrimination (ID)
and \textit{online} cluster discrimination (CD) designs 
in the \abbrname model.
For models trained without cross-cluster ID, 
all the memory banks were concatenated as 
the contrastive set for every sample (Eq.~\eqref{eq:memory}), 
whilst the cluster assignments $\tilde{\bm{p}}$ 
yielded by the semantic memory (Eq.~\eqref{eq:semantic_memory_prediction}) 
was used for pseudo labels updating 
if learned without online CD.
Instance contrastive learning was considered as the baseline
without both the ID and CD components of SCL.
As shown in Fig.~\ref{fig:ablation},
the models trained without cross-cluster ID or online CD 
can always surpass instance contrastive learning with remarkable margins,
which demonstrates their effectiveness as individual components. 
By jointly learning with both, 
\abbrname always produced superior performances
which indicates the mutual benefits of representation learning 
and decision boundaries reasoning.

\begin{figure}[ht]
\begin{minipage}[t]{0.48\linewidth}
\centering
\includegraphics[width=\textwidth]{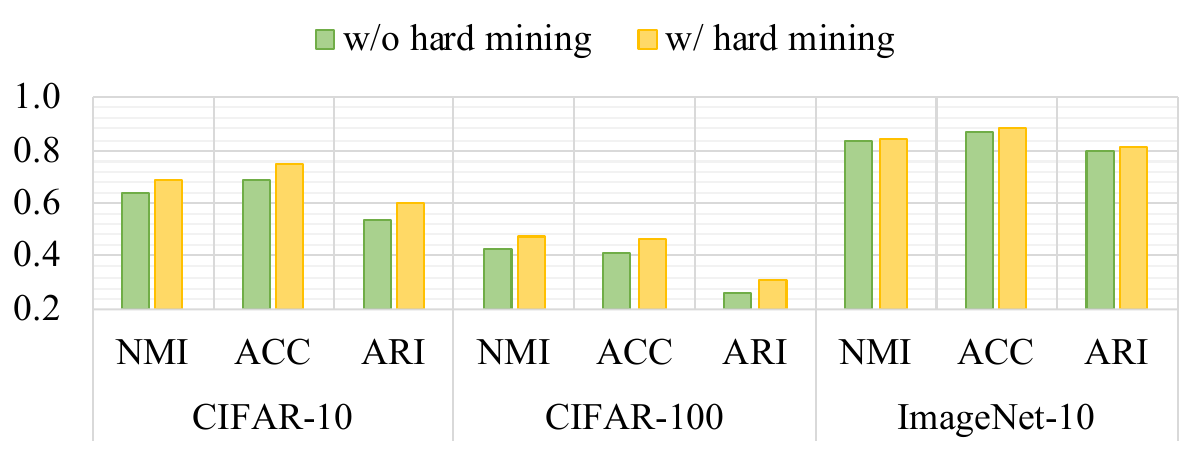}
\caption{An ablation study on the hard sample mining strategy.}
\label{fig:reweight}
\end{minipage}
\hfill
\begin{minipage}[t]{0.48\linewidth}
\centering
\includegraphics[width=\textwidth]{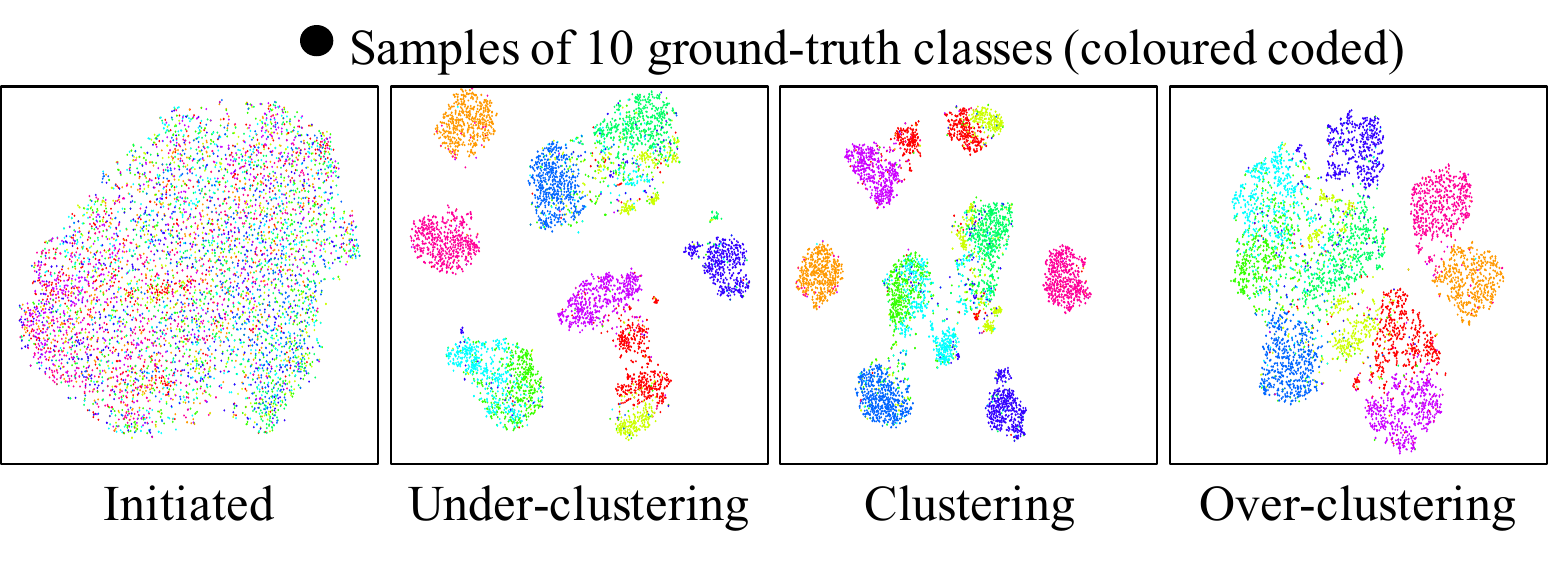}
\caption{Feature visualisation for images on CIFAR-10
using t-SNE~\cite{maaten2008tsne}.}
\label{fig:tsne}
\end{minipage}
\end{figure}

\paragraph{Hard sample mining.}
To emphasise the hard samples in model learning,
we re-weighted the samples within the same mini-batches
according to their assignment stability (Eq.~\eqref{eq:reweight_loss}).
To study the effectiveness of this design,
we replaced it by averaging their losses
as in conventional batch-wise training.
According to Fig.~\ref{fig:reweight},
the learned clusters show higher consistency
with the ground-truth classes
when training with the re-weighting strategy.
This demonstrates the importance of higtlighting hard samples
with ambiguous semantic meanings 
to improve the model's class discrimination capability.

\paragraph{Feature visualisation.}
To better understand model effectiveness, we visualise some
sample representations from a randomly initialised model and 
those learned from different clustering tasks
with different cluster numbers
(under-, over- and clustering)
on CIFAR-10.
Fig.~\ref{fig:tsne} shows that
the initial states of the feature spaces were chaotic,
which would certainly lead to error-propagation
if trained by estimated assignments
in a conventional supervised learning process.
Whilst the `over-clustering' task resulted in 
less within-cluster compactness than `clustering' and `under-clustering',
`under-clustering' yielded less separable clusters.
By jointly training on all three tasks,
\abbrname explores visual similarity in multiple granularities
and learns clusters according to their consensus,
hence, more robust to visual ambiguity.

\begin{wrapfigure}{r}{0.5\textwidth}
\includegraphics[width=\linewidth]{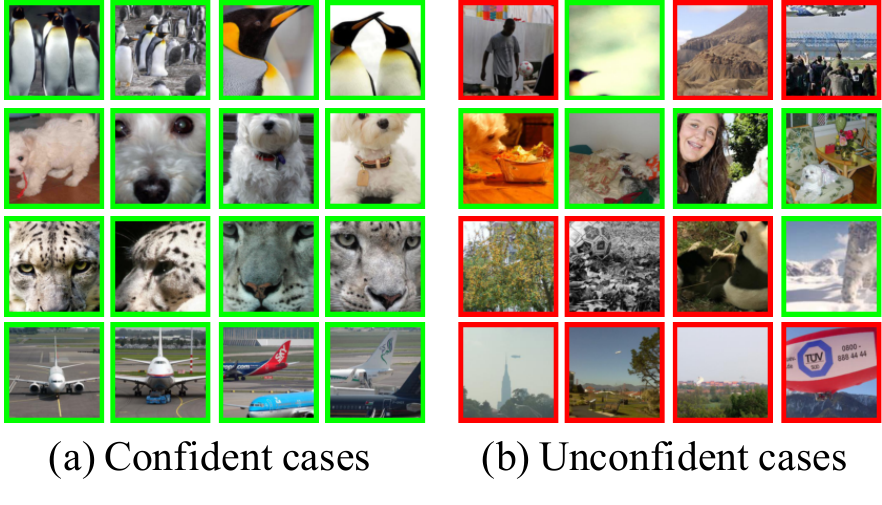}
\caption{Examples from ImageNet-10.
Per class each row:
(a) top-4 `confident' and
(b) bottom-4 `unconfident' cases
\wrt assignment probabilities.
Samples in green boxes are assigned to the correct classes
while those with red boxes are failed cases.}
\label{fig:cases}
\end{wrapfigure}

\paragraph{Visual case examples.}
Fig.~\ref{fig:cases} shows two groups of 
image examples from ImageNet-10, 
with the highest/lowest probabilities (confident/unconfident) 
for being in a cluster shown in each row.
It is evident that the assignment confidence yielded by \abbrname
is well-aligned with the correctness of model predictions.
This means that the most confident label interpretations of the
learned clusters are also more likely in agreement with the ground-truth
categories, \ie semantic plausibility is consistent with the model
prediction confidence.
Most of the failed cases are due to images being significantly
dominated by background.
This suggests that it is challenging
for unsupervised learning to identify correctly the relevant
focus of attention in a visual context.

\section{Conclusion}
\label{sec:conclusion}
In this work,
we proposed a novel \textit{\fullname} (\abbrname)
method for high-level semantic understanding of visual data 
without learning from manual labels.
The \abbrname model
addresses the fundamental limitation of instance contrastive learning
by imposing the cluster structure into the 
unlabelled training data
so to jointly learn discriminative visual feature representations 
and reason about cluster decision boundaries
while avoiding the inherent contradiction
between their learning objectives.
By learning visual features with high robustness to temporal
(intermediate) cluster assignments in the course of model training,
\abbrname mitigates the common error-propagation problem
of contemporary deep clustering techniques. 
Moreover,
by exploring semantic relations from contrastive visual similarity,
the clusters yielded by \abbrname 
encode unique and consistent visual characteristics. 
Hence, 
\abbrname is semantically more plausible.
Experiments on six object recognition datasets
show the \abbrname's superiority
over the state-of-the-art deep clustering
and instance contrastive models.

\section*{Acknowledgements}
This work was supported by the China Scholarship Council, Vision Semantics Limited, 
the Alan Turing Institute Turing Fellowship.

\bibliography{press,egbib}
\end{document}